\pdfoutput=1

\documentclass[11pt]{article}

\usepackage{EMNLP2023}
\usepackage{graphicx}

\usepackage{times}
\usepackage{latexsym}

\usepackage[T1]{fontenc}

\usepackage[utf8]{inputenc}

\usepackage{microtype}

\usepackage{inconsolata}

\title{Cogs in a Machine, Doing What They’re Meant to Do \\– The AMI Submission to the WMT24 General Translation Task}

 \author{Atli Jasonarson, Hinrik Hafsteinsson, Bjarki Ármannsson, Steinþór Steingrímsson \\
         The Árni Magnússon Institute for Icelandic Studies\\
         Reykjavík, Iceland\\\texttt{atli.jasonarson,hinrik.hafsteinsson,bjarki.armannsson,}\\\texttt{steinthor.steingrimsson@arnastofnun.is}}

\begin{document}

\maketitle

\begin{abstract}

This paper presents the submission of the Árni Magnusson Institute's team to the WMT24 General translation task. We work on the English$\rightarrow$Icelandic translation direction. Our system comprises four translation models and a grammar correction model. For training our models we carefully curate our datasets, aggressively filtering out sentence pairs that may detrimentally affect the quality of our system's output. Some of our data are collected from human translations and some are synthetically generated. A part of the synthetic data is generated using an LLM, and we find that it increases the translation capability of our system significantly.
\end{abstract}

\section{Introduction}
We describe our submission to the 2024 WMT general translation task. 
Large Language Models (LLMs) have become near-ubiquitous in the field of Natural Language Processing (NLP) in the last couple of years. They have shown remarkable translation capabilities (see e.g. \citeauthor{xu2024a},~\citeyear{xu2024a}), but require significantly larger computational resources than previous neural MT (NMT) models, both for training and inference. Most openly available LLMs are primarily trained on English texts and may therefore need further training in order to be able to translate from or into less-resourced languages, such as Icelandic. 

The ALMA models \cite{xu2024a} are LLM-based translation models, built on LLaMA-2. They have been trained to translate ten directions, including English$\leftrightarrow$Icelandic. We explore the capabilities of some of these models, the 7B and 13B parameter versions of ALMA-R \cite{xu2024contrastivepreferenceoptimizationpushing}, and find that they generate very competitive translations as measured against the English--Icelandic WMT21 test sets \cite{akhbardeh-etal-2021-findings}, especially from Icelandic into English. Unfortunately, using our settings the translation speed was quite slow (approximately one sentence per second) on an NVIDIA A100 GPU card. 

We are interested in building faster models so we use the more traditional encoder-decoder Transformer architecture described in \citet{vaswani2017}. We collect all parallel data available to us for our language pair, generate additional synthetic pairs using the ALMA-R 13B parameter model and apply iterative back-translation using our own models.  We apply filters to remove sentence pairs that may have detrimental effects on the models output.





We train four Transformer models\footnote{Models available at \url{https://huggingface.co/arnastofnun}.} of varying sizes and let each model generate five translation candidates. A spelling and grammar checking model is then applied to the translations to generate ``corrected'' versions of the sentences. Finally the best candidate is selected from the pool of translations, corrected or not, using a reranking model.

We evaluate our models and approaches on the WMT21 test set for English$\rightarrow$Icelandic.


\section{Related Work}
We only submit a system for the English$\rightarrow$Icelandic translation direction. This language pair was previously one of the pairs for the WMT General Translation shared task in 2021 but prior to that, limited work had been published on MT for Icelandic. \citet{brandt-etal-2011-apertium} describe a rule-based system for translating Icelandic$\rightarrow$English, based on Apertium \cite{apertium2011}. \citet{DBLP:conf/tsd/JonssonSSSL20} was the first published work describing SMT and NMT for Icelandic. Since 2021 the WMT21 evaluation data, as well as various parallel corpora projects, have made it more accessible to train and evaluate MT systems translating to or from Icelandic, and with that the language has been included in various research projects. We believe this is an indicator of the importance of evaluation campaigns, such as the ones run in association with the WMT conferences, for less prominent languages.

Our approach uses an ensemble of four different translation models and a reranking model to select the best candidate. This is a common approach, motivated by the intuition that different systems may have different strengths. In recent work, \citet{toraletal2023} use this approach in their experiments with literary translations. In their work on bidirectional reranking, \citet{Imamura2017EnsembleAR} discuss reranking and ensembling for MT in some detail. Examples from the period of statistical MT include the work of \citet{olteanu-etal-2006-language} and \citet{4430102}, describing language model-based reranking on hypotheses generated by phrase-based SMT systems.

\section{Data Selection and Filtering}

Various parallel data are available for the English--Icelandic language pair. ParIce \cite{barkarson-steingrimsson-2019-compiling} is partly a collection of parallel corpora available elsewhere, which has been realigned and refiltered, and partly data compiled for that project, the largest source being regulatory texts published in relation with the European Economic Area (EEA) agreement. Data for the English--Icelandic language pair were collected within the Paracrawl project \cite{banon-etal-2020-paracrawl}, CCMatrix \cite{schwenk-etal-2021-ccmatrix}, MaCoCu \cite{non-etal-2022-macocu} and HPLT \cite{aulamo-etal-2023-hplt}. Data for the language pair are also available from multiple smaller datasets distributed on OPUS \cite{tiedemann-thottingal-2020-opus}. We utilize all these datasets in training our models.

We also use synthetic data: Backtranslations made available by \citet{20.500.12537/260}, translations generated using the ALMA-R 13B parameter model and backtranslations generated by our trained models. We describe these in more detail in Section \ref{sec:synthetic}.

\citet{khayrallah-koehn-2018-impact} show that incorrect translations, untranslated target text, misalignments, and other noisy segments in training data can have a detrimental effect on the quality of translations generated by NMT systems trained on that data. By filtering our training data rather aggressively, we try to minimize such noise.

\subsection{ParIce}
\label{sec:parice}
Even though care has been taken to realign and refilter data for the ParIce corpus, \citet{steingrimsson-etal-2023-filtering} show that it still contains noise, such as misalignments and mistranslations, that may be detrimental when training NMT systems. They refilter the data using a combination of approaches: Shallow filters based on simple heuristics, by using Bicleaner \cite{sanchez-cartagena-etal-2018-prompsits,prompsit:2020:EAMT} and by employing classifiers (support vector machine-based ones \cite{cortes1995support} had the best outcome) with a combination of scoring mechanisms, including LASER \cite{Artetxe2018MassivelyMS}, LaBSE \cite{feng-etal-2022-language}, NMTScore \cite{https://doi.org/10.48550/arxiv.2204.13692} using the M2M100 multilingual translation model \cite{10.5555/3546258.3546365}, and WAScore, a word alignment-based score devised to measure word-level parallelism, introduced in \citet{steingrimsson-etal-2021-effective}. In \citet{Steingrimsson2023Phd} these data are processed further by realigning the EEA texts in the ParIce corpus using SentAlign \cite{steingrimsson-etal-2023-sentalign}. 

As the basis for our training we use the ParIce dataset, processed as described above, as well as parallel data extracted from Wikipedia using the comparable corpora mining approach described in \citep{steingrimsson-etal-2021-effective} and sentence pairs extracted from version 9 of Paracrawl using the filtering approaches described above and in \citet{steingrimsson-etal-2023-filtering}.

\begin{figure*}
    \centering\includegraphics[width=1\linewidth]{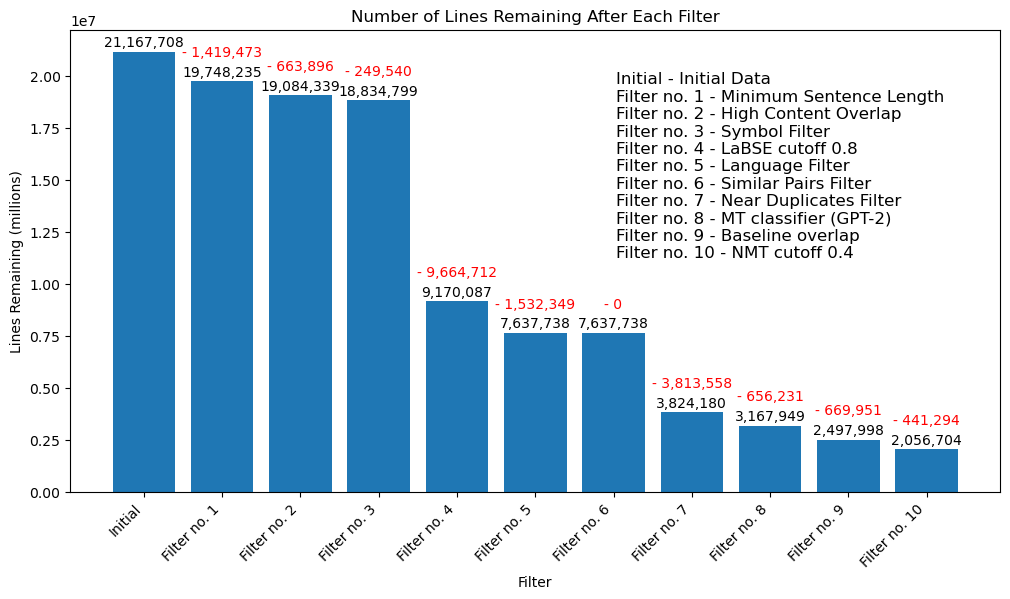}
    \caption{Each filtering step's effect on OPUS dataset size}
    \label{fig:filter-stats}
\end{figure*}

\subsection{Filtering the OPUS Datasets}\label{sec:opus}
An overview of the data for Icelandic-English parallel texts sourced from the OPUS catalog is provided in Appendix \ref{sec:appendix_opus}. This data, accounting for redundant sentence pairs, amounts to 21.167.708\footnote{This applies to the state of the OPUS catalog at the time of development, i.e., April 2024.} sentence pairs. At face value, this is a substantial amount of available data. However, the quality of these parallel texts is not reliable, with noisy and incorrect pairs being prevalent throughout most individual datasets in the catalog. To remedy this, and thus ensure that the data sourced via OPUS can be used effectively in our project, we applied an aggressive, sequential filtering process, with the goal of whittling away the majority of the low-quality sentence pairs.

Our sequential filtering process consists of ten individual steps, most of which only remove sentences from the data without modifying the content of other sentences. The process is \textit{sequential}, in that the input of a filtering step is the output of the previous filtering step. Furthermore, the order of these steps is decided to ensure optimal processing time of the filters so that computationally heavy filtering steps process the least amount of data, which minimizes run time. For a detailed overview of each filtering step, see Appendix \ref{sec:appendix_filter}.



The effects of each filtering step on the data amount is shown in Fig. \ref{fig:filter-stats}.
To ensure that our filtering methods affected our implementation positively, we intermittently added the output of the filtering process to our training pipeline and evaluated the performance. In particular, we used this approach to dial in the optimal LaBSE and NMT score cutoffs in our filters. 

The final output of our filtering process produces a relatively high-quality data set of 2.056.704 English-Icelandic sentence pairs (roughly 9.71\% of the original 21.167.708 raw sentence pairs sourced from the OPUS catalog), which we then add to our training data.

\subsection{Synthetic Data}
\label{sec:synthetic}
The dataset made available by \citet{20.500.12537/260} contains translations from Europarl, Newscrawl, Wikipedia and the IGC. We perform a filtering step similar to the one used applied on the OPUS data, consisting of a length filter, removing all sentences that have fewer than four word tokens and more than 150, an overlap filter, removing all sentence pairs that share 40\% or more of word tokens, and a symbol filter removing all sentence pairs where more than 20\% of characters in one of the sentences is non-alphabetical. Furthermore we use two scoring mechanisms for filtering, LaBSE, using a score threshold of 0.8, and NMTScore with a threshold of 0.4. These scores are selected based on the evaluation in \cite{steingrimsson-etal-2023-filtering}. After filtering, we are left with 4.4M sentence pairs from this dataset.

We use the 13B parameter ALMA-R model to translate English sentences from Newscrawl to Icelandic and Icelandic texts from the Icelandic Gigaword Corpus (IGC) \cite{steingrimsson_2018} to English. The Icelandic texts are sampled from three different subcorpora of the IGC, comprising news, scholarly journals, and literary texts. For each source sentence we generate five translations and use LaBSE to select the two best ones, granted that they exceed a threshold of a LaBSE score of 0.8 and pass through the three shallow filters described above: length, overlap and symbol filters. Our final set contains 8.9M sentence pairs translated from Icelandic to English and 700K sentence pairs translated from English to Icelandic.

Finally, we do iterative back-translation. We use the same training data as described above to train models to translate texts from the IGC to English. For the back-translations we use Transformer$_\mathrm{BIG}$ models \citep{vaswani2017}, as described in Table \ref{tab:modeldesc}. We use the same approach as before, generate five translations for each sentence and use LaBSE to select the two best ones, as long as they exceed the threshold of 0.8 and are not filtered out by the other filters. We do two iterations of translating and training models in both translation directions using backtranslated data. This results in a total of approximately 60M sentence pairs.

\begin{table}
\centering
\begin{tabular}{lccccc}
\hline
\textbf{model} & \textbf{$d_{model}$} & \textbf{$d_{ff}$} & \textbf{$h$} & \textbf{$N_{enc}$} & \textbf{$N_{dec}$}\\
\hline
$Base$&512&2048&8&6&6\\
$Base_{deep}$&512&2048&8&36&12\\
$Big$&1024&4096&16&6&6\\
$Big_{deep}$&1024&4096&16&36&12\\
\hline
\end{tabular}
\caption{Model dimensions, heads and number of layers.}
\label{tab:modeldesc}
\end{table}

\begin{table}[b!]
\centering
\begin{tabular}{lc}
\hline
\textbf{Dataset} & \textbf{chrF}\\
\hline
Baseline&50.4\\
Baseline+lexicon&50.4\\
Baseline+OPUS&53.7\\
Baseline+Jónsson&53.5\\
Baseline+Jónsson+SMT&53.2\\
Baseline+Jónsson+ALMA&54.7\\
Baseline+Jónsson+ALMA+OPUS&55.1\\
Baseline+Jónsson+ALMA+OPUS+BT1&56.4\\
Baseline+Jónsson+ALMA+OPUS+BT2&56.8\\
\hline
\end{tabular}
\caption{The table shows that when most of the datasets in our experiments are added to the training data the quality, as measured by chrF, increases. Exceptions to that are the experiments with adding token-pairs from an English-Icelandic lexicon and with using backtranslations generated by an SMT system. These two datasets are therefore not used in our final systems.}
\label{tab:dataselection}
\end{table}

\subsection{Other Data}
\label{sec:other_data}
To decide which datasets to use, we trained Transformer$_\mathrm{BASE}$ models as described in \citet{vaswani2017} and evaluated the models using the test set from WMT21. We started by training a baseline system using the dataset described in Section \ref{sec:parice}. We then added different datasets to the baseline data, trained new systems and evaluated them. If the new dataset seemed to improve the output we used that for our final system. In addition to previously described datasets we tried generating backtranslations using SMT and to add data from a bilingual lexicon using token-pair training as described by \citet{jones-etal-2023-gatitos}. Table \ref{tab:dataselection} shows chrF scores \cite{popovic-2015-chrf} for our different experiments.

The total number of sentence pairs used for training is shown in Table \ref{tab:data}

\begin{table}
\centering
\begin{tabular}{lc}
\hline
\textbf{Dataset} & \textbf{Sentence Pairs}\\
\hline
Base&2,277,023\\
OPUS-filtered&2,056,704\\
Miðeind-BT&2,559,806\\
Miðeind-FT&1,837,945\\
ALMA-BT&8,927,720\\
ALMA-FT&700,253\\
IGC-BT-1&27,794,398\\
IGC-BT-2&33,465,175\\
\hline
\end{tabular}
\caption{Datasets used for training and number of sentence pairs in each dataset.}
\label{tab:data}
\end{table}

\section{System Description}

Our motivation for using multiple models is twofold: First, we want to use models that are computationally inexpensive to run and so we train models that can run on one consumer grade GPU. Second, systems of different sizes may have complementary strengths and so training multiple systems and reranking the results may give us better results than any one model.

We train four encoder-decoder Transformer models, all of which play a part in the translation pipeline. Two of the models follow the exact architecture described in~\citet{vaswani2017}, i.e. the `base' and `big' versions of the original Transformer model, while the other two are deeper, using 36 encoder layers and 12 decoder layers instead of six. The difference between the four models is shown in Table \ref{tab:modeldesc}.

The outputs from the translation models undergo two post-processing steps. First, they are run through a grammatical error correction model, a version of the byte-level sequence-to-sequence model ByT5~\cite{xue2022byt5tokenfreefuturepretrained} that has been fine-tuned by ~\citet{ingólfsdóttir2023bytelevelgrammaticalerrorcorrection} to correct spelling errors in Icelandic as well as handling more complex grammatical, semantic and stylistic issues. Second, we fix punctuation errors which translation models are prone to making when translating into Icelandic (mostly to do with quotation marks, which are different in Icelandic and English) as well as some that might be unique to our system, such as their incapability to translate emojis. As the grammatical error correction model proved too aggressive for our purposes, merging and splitting some sentences, normalizing informal language usage and hashtags, etc., we also revert some of the changes it introduced. 

Using the WMT21 test set we experiment with an ensemble approach, using \textsc{CometKiwi-da-22}~\cite{rei-etal-2022-cometkiwi} to select the best sentence out of 20 hypotheses made by the four models (each model generates five hypotheses using beam search with beam size 12). This raises the chrF score to 58.3 for our evaluation set. On top of this we add the spelling and grammar error correction, which gives us a very modest increase in quality as measured by chrF, shown in Table \ref{tab:modelscores}.

We investigate whether the \textsc{CometKiwi-da-22} model prefers the output from some of the translation models over the others. Table \ref{tab:modelselection} shows which translation models generated the translations ultimately chosen by the scoring model when experimenting on the WMT21 evaluation set of 1000 sentences. While translations by the deeper model are more likely to be selected, it is evident that all models are contributing, with the final selection containing 753 translation generated by only one model, and of these all models contribute over 150 translations each. 247 of the selected translations were generated by more than one model (non-unique translations). An ensemble approach thus seems to be likely to improve overall translation quality.

\begin{table}[t]
\centering
\begin{tabular}{lc}
\hline
\textbf{model} & chrF\\
\hline
$Base$&56.8\\
$Base_{deep}$&57.1\\
$Big$&57.7\\
$Big_{deep}$&57.7\\
Ensemble+\textsc{CometKiwi}&58.3\\
Ensemble+error correction&\\\hspace{3ex}+\textsc{CometKiwi}&\textbf{58.4}\\
\hline
\hline
ALMA-R 7B&52.2\\
ALMA-R 13B&53.4\\
\hline
\end{tabular}
\caption{chrF scores for each of our models, compared with scores for the model ensembles and for the ALMA-R models. The scores are calculated on the WMT21 evaluation set.}
\label{tab:modelscores}
\end{table}

\subsection{The pipeline}
Basing our system on the most succesful approach in our experiments, our translation pipeline consists of three steps: First, using each of our four models, we generate five translation hypotheses using beam search for all source paragraphs, resulting in a total of 20 candidates. 

Furthermore, each paragraph is segmented into sentences, $s_1, \ldots, s_n$. For each sentence, every model produces five hypotheses. These hypotheses are evaluated using \textsc{CometKiwi-da-22}, and the highest-scoring hypothesis is selected for each sentence. The selected hypotheses are concatenated to form a new paragraph.
Finally, a single paragraph is created by combining the best translation of each sentence, leaving us with 25 translation candidates.

Each of these candidates is then corrected with regard to grammar, spelling and style using the ByT5 model described above. 

These two steps, translating the source text and correcting the translations, result in a total of 50 translation candidates. In order to find the best candidate we use \textsc{CometKiwi-da-22} to score all candidates. The highest scoring one is the selected translation of our system.

\begin{table}[t]
\centering
\begin{tabular}{lcc}
\hline
\textbf{model} & \textbf{Selected} & \textbf{Unique}\\
\hline
$Base$&293&158\\
$Base_{deep}$&347&186\\
$Big$&287&163\\
$Big_{deep}$&419&246\\
\hline
\end{tabular}
\caption{The number of sentences generated by each model selected for the final output when translating the WMT21 test set. }
\label{tab:modelselection}
\end{table}






\begin{table*}[]
\centering
\begin{tabular}{lcccc}

\bf System Name  & \bf Type  & \bf AutoRank $\downarrow$ & \bf MetricX $\downarrow$ & \bf CometKiwi $\uparrow$ \\
Unbabel-Tower70B & Closed & 1.0 & 2.5 & 0.740 \\
Claude-3.5 & Closed & 2.3 & 3.6 & 0.697 \\
Dubformer &Closed &  2.5 & 3.4 & 0.685 \\
IKUN & Open & 3.2 & 4.3 & 0.666 \\
GPT-4 & Closed & 3.4 & 4.7 & 0.673 \\
\textbf{AMI} & \textbf{Open} & \textbf{3.7} & \textbf{4.9} & \textbf{0.663} \\
IKUN-C & Constrained & 3.7 & 4.9 & 0.657 \\
TranssionMT & Closed & 4.2 & 5.5 & 0.653 \\
ONLINE-B & Closed & 4.2 & 5.5 & 0.652 \\
IOL-Research & Open & 4.3 & 5.7 & 0.655 \\
ONLINE-A & Closed & 5.5 & 6.4 & 0.603 \\
Llama3-70B & Open & 6.7 & 8.0 & 0.586 \\
ONLINE-G & Closed & 6.9 & 7.9 & 0.573 \\
CommandR-plus & Closed & 9.8 & 10.6 & 0.487 \\
Mistral-Large & Closed & 10.4 & 10.9 & 0.465 \\
Aya23 & Open & 15.2 & 14.9 & 0.311 \\
Phi-3-Medium & Closed & 16.2 & 15.7 & 0.278 \\
ONLINE-W & Closed & 18.1 & 19.5 & 0.296 \\
TSU-HITs & Constrained & 19.2 & 18.4 & 0.192 \\
CycleL & Constrained & 21.0 & 20.2 & 0.148 \\
\end{tabular}
\caption{Preliminary WMT24 General MT automatic ranking for English-Icelandic. Our system is in bold.}
\label{tab:results}
\end{table*}

\section{Results}
We evaluate our system on the test data from WMT21. As expected, the bigger models perform better, but the best results are achieved by selecting translations from an ensemble of differently trained Transformer models. We use \textsc{CometKiwi-da-22} to select the best translation out of 20 hypotheses made by the four models, five hypotheses by each using beam search with beam size 12. This raises the chrF score to 58.3 and when we add error correction on top, the score is slightly higher, 58.4, as shown in Table \ref{tab:modelscores}.


In the WMT24 general translation task, systems were evaluated using two automatic metrics, MetricX-23-XL \cite{juraska-etal-2023-metricx} and \textsc{CometKiwi-DA-XL} \cite{rei-etal-2023-scaling}, as well as by human evaluation. According to the automatic metrics, reported in \citet{kocmi2024preliminarywmt24rankinggeneral}, our model is competitive among the open systems, although four closed systems achieve better scores. Results for the automatic metrics are shown in Table \ref{tab:results}.

\section{Conclusions and Future Work}
We show that while Large Language Models have become nearly ubiquitous in Natural Language Processing, traditional encoder-decoder Transformer models remain a viable approach to machine translation, particularly when computational efficiency is a priority. 

Nevertheless, our findings also reveal that integrating LLMs can be advantageous during the training process. Specifically, \mbox{ALMA-R} 13B proved to be an important part of our training pipeline, as the synthetic data it generated increased the quality of our translation systems. 

Furthermore, our results indicate that while more training data usually result in a better translation system, low-quality data, such as the backtranslations generated with an SMT system, can have a detrimental impact on performance. Similarly, our experiments with a bilingual lexicon using token-pair training negatively affected the system's output. This may be due to a variety of reasons. Our SMT system could probably be improved as well as our approach to include data from a bilingual lexicon in the training data. This warrants further investigation.



Our filtering method, as described in Sections \ref{sec:opus}, \ref{sec:synthetic} and Appendix \ref{sec:appendix_filter}, has proven effective, even though it may be argued that it is still somewhat crude and more work into minimizing the loss of useful sentence pairs and more effectively remove detrimental sentence pairs would very likely improve the training data and in turn the translation models. For example, while we use LaBSE, LASER and NMT to evaluate sentence pairs, we apply individual cutoff values for each score. A better approach could entail using a classifier to combine all metrics for an optimal result.

Although currently impractical at production-scale, genetic algorithms, as shown by \citet{jon-bojar-2023-breeding} and \citet{jon-etal-2023-cuni}, show promising results in generating translation candidates. Given larger computational resources, similar approaches might prove useful and await future study.




\bibliography{emnlp2023}

\begin{thebibliography}{46}
\expandafter\ifx\csname natexlab\endcsname\relax\def\natexlab#1{#1}\fi

\bibitem[{Akhbardeh et~al.(2021)Akhbardeh, Arkhangorodsky, Biesialska, Bojar, Chatterjee, Chaudhary, Costa-jussa, Espa{\~n}a-Bonet, Fan, Federmann, Freitag, Graham, Grundkiewicz, Haddow, Harter, Heafield, Homan, Huck, Amponsah-Kaakyire, Kasai, Khashabi, Knight, Kocmi, Koehn, Lourie, Monz, Morishita, Nagata, Nagesh, Nakazawa, Negri, Pal, Tapo, Turchi, Vydrin, and Zampieri}]{akhbardeh-etal-2021-findings}
Farhad Akhbardeh, Arkady Arkhangorodsky, Magdalena Biesialska, Ond{\v{r}}ej Bojar, Rajen Chatterjee, Vishrav Chaudhary, Marta~R. Costa-jussa, Cristina Espa{\~n}a-Bonet, Angela Fan, Christian Federmann, Markus Freitag, Yvette Graham, Roman Grundkiewicz, Barry Haddow, Leonie Harter, Kenneth Heafield, Christopher Homan, Matthias Huck, Kwabena Amponsah-Kaakyire, Jungo Kasai, Daniel Khashabi, Kevin Knight, Tom Kocmi, Philipp Koehn, Nicholas Lourie, Christof Monz, Makoto Morishita, Masaaki Nagata, Ajay Nagesh, Toshiaki Nakazawa, Matteo Negri, Santanu Pal, Allahsera~Auguste Tapo, Marco Turchi, Valentin Vydrin, and Marcos Zampieri. 2021.
\newblock \href {https://aclanthology.org/2021.wmt-1.1} {Findings of the 2021 conference on machine translation ({WMT}21)}.
\newblock In \emph{Proceedings of the Sixth Conference on Machine Translation}, pages 1--88, Online. Association for Computational Linguistics.

\bibitem[{Artetxe and Schwenk(2019)}]{Artetxe2018MassivelyMS}
Mikel Artetxe and Holger Schwenk. 2019.
\newblock {Massively Multilingual Sentence Embeddings for Zero-Shot Cross-Lingual Transfer and Beyond}.
\newblock \emph{Transactions of the Association for Computational Linguistics}, 7:597–610.

\bibitem[{Aulamo et~al.(2023)Aulamo, Bogoychev, Ji, Nail, Ram{\'\i}rez-S{\'a}nchez, Tiedemann, van~der Linde, and Zaragoza}]{aulamo-etal-2023-hplt}
Mikko Aulamo, Nikolay Bogoychev, Shaoxiong Ji, Graeme Nail, Gema Ram{\'\i}rez-S{\'a}nchez, J{\"o}rg Tiedemann, Jelmer van~der Linde, and Jaume Zaragoza. 2023.
\newblock \href {https://aclanthology.org/2023.eamt-1.61} {{HPLT}: High performance language technologies}.
\newblock In \emph{Proceedings of the 24th Annual Conference of the European Association for Machine Translation}, pages 517--518, Tampere, Finland. European Association for Machine Translation.

\bibitem[{Ba{\~n}{\'o}n et~al.(2020)Ba{\~n}{\'o}n, Chen, Haddow, Heafield, Hoang, Espl{\`a}-Gomis, Forcada, Kamran, Kirefu, Koehn, Ortiz~Rojas, Pla~Sempere, Ram{\'\i}rez-S{\'a}nchez, Sarr{\'\i}as, Strelec, Thompson, Waites, Wiggins, and Zaragoza}]{banon-etal-2020-paracrawl}
Marta Ba{\~n}{\'o}n, Pinzhen Chen, Barry Haddow, Kenneth Heafield, Hieu Hoang, Miquel Espl{\`a}-Gomis, Mikel~L. Forcada, Amir Kamran, Faheem Kirefu, Philipp Koehn, Sergio Ortiz~Rojas, Leopoldo Pla~Sempere, Gema Ram{\'\i}rez-S{\'a}nchez, Elsa Sarr{\'\i}as, Marek Strelec, Brian Thompson, William Waites, Dion Wiggins, and Jaume Zaragoza. 2020.
\newblock \href {https://doi.org/10.18653/v1/2020.acl-main.417} {{ParaCrawl: Web-Scale Acquisition of Parallel Corpora}}.
\newblock In \emph{Proceedings of the 58th Annual Meeting of the Association for Computational Linguistics}, pages 4555--4567, Online. Association for Computational Linguistics.

\bibitem[{Ba{\~n}{\'o}n et~al.(2022)Ba{\~n}{\'o}n, Espl{\`a}-Gomis, Forcada, Garc{\'\i}a-Romero, Kuzman, Ljube{\v{s}}i{\'c}, van Noord, Sempere, Ram{\'\i}rez-S{\'a}nchez, Rupnik, Suchomel, Toral, van~der Werff, and Zaragoza}]{non-etal-2022-macocu}
Marta Ba{\~n}{\'o}n, Miquel Espl{\`a}-Gomis, Mikel~L. Forcada, Cristian Garc{\'\i}a-Romero, Taja Kuzman, Nikola Ljube{\v{s}}i{\'c}, Rik van Noord, Leopoldo~Pla Sempere, Gema Ram{\'\i}rez-S{\'a}nchez, Peter Rupnik, V{\'\i}t Suchomel, Antonio Toral, Tobias van~der Werff, and Jaume Zaragoza. 2022.
\newblock \href {https://aclanthology.org/2022.eamt-1.41} {{M}a{C}o{C}u: Massive collection and curation of monolingual and bilingual data: focus on under-resourced languages}.
\newblock In \emph{Proceedings of the 23rd Annual Conference of the European Association for Machine Translation}, pages 303--304, Ghent, Belgium. European Association for Machine Translation.

\bibitem[{Barkarson and Steingr{\'\i}msson(2019)}]{barkarson-steingrimsson-2019-compiling}
Starka{\dh}ur Barkarson and Stein{\th}{\'o}r Steingr{\'\i}msson. 2019.
\newblock \href {https://aclanthology.org/W19-6115} {{Compiling and Filtering {P}ar{I}ce: An {E}nglish-{I}celandic Parallel Corpus}}.
\newblock In \emph{Proceedings of the 22nd Nordic Conference on Computational Linguistics}, pages 140--145, Turku, Finland. Link{\"o}ping University Electronic Press.

\bibitem[{Brandt et~al.(2011)Brandt, Loftsson, Sigur{\th}{\'o}rsson, and Tyers}]{brandt-etal-2011-apertium}
Martha~D{\'\i}s Brandt, Hrafh Loftsson, Hlynur Sigur{\th}{\'o}rsson, and Francis~M. Tyers. 2011.
\newblock \href {https://aclanthology.org/2011.eamt-1.30} {{Apertium-IceNLP: A rule-based Icelandic to English machine translation system}}.
\newblock In \emph{Proceedings of the 15th Annual conference of the European Association for Machine Translation}, pages 217--224, Leuven, Belgium. European Association for Machine Translation.

\bibitem[{Cortes and Vapnik(1995)}]{cortes1995support}
Corinna Cortes and Vladimir Vapnik. 1995.
\newblock Support-vector networks.
\newblock \emph{Machine learning}, 20(3):273--297.

\bibitem[{Fan et~al.(2021)Fan, Bhosale, Schwenk, Ma, El-Kishky, Goyal, Baines, Celebi, Wenzek, Chaudhary, Goyal, Birch, Liptchinsky, Edunov, Grave, Auli, and Joulin}]{10.5555/3546258.3546365}
Angela Fan, Shruti Bhosale, Holger Schwenk, Zhiyi Ma, Ahmed El-Kishky, Siddharth Goyal, Mandeep Baines, Onur Celebi, Guillaume Wenzek, Vishrav Chaudhary, Naman Goyal, Tom Birch, Vitaliy Liptchinsky, Sergey Edunov, Edouard Grave, Michael Auli, and Armand Joulin. 2021.
\newblock {Beyond English-Centric Multilingual Machine Translation}.
\newblock \emph{J. Mach. Learn. Res.}, 22(1).

\bibitem[{Feng et~al.(2022)Feng, Yang, Cer, Arivazhagan, and Wang}]{feng-etal-2022-language}
Fangxiaoyu Feng, Yinfei Yang, Daniel Cer, Naveen Arivazhagan, and Wei Wang. 2022.
\newblock \href {https://doi.org/10.18653/v1/2022.acl-long.62} {{Language-agnostic {BERT} Sentence Embedding}}.
\newblock In \emph{Proceedings of the 60th Annual Meeting of the Association for Computational Linguistics (Volume 1: Long Papers)}, pages 878--891, Dublin, Ireland. Association for Computational Linguistics.

\bibitem[{Forcada et~al.(2011)Forcada, Ginestí-Rosell, Nordfalk, O'Regan, Ortiz-Rojas, Pérez-Ortiz, Sánchez-Martínez, Ramírez-Sánchez, and Tyers}]{apertium2011}
Mikel Forcada, Mireia Ginestí-Rosell, Jacob Nordfalk, Jim O'Regan, Sergio Ortiz-Rojas, Juan Pérez-Ortiz, Felipe Sánchez-Martínez, Gema Ramírez-Sánchez, and Francis Tyers. 2011.
\newblock \href {https://doi.org/10.1007/s10590-011-9090-0} {Apertium: A free/open-source platform for rule-based machine translation}.
\newblock \emph{Machine Translation}, 25:127--144.

\bibitem[{Imamura and Sumita(2017)}]{Imamura2017EnsembleAR}
Kenji Imamura and Eiichiro Sumita. 2017.
\newblock \href {https://aclanthology.org/W17-5711} {Ensemble and reranking: Using multiple models in the {NICT}-2 neural machine translation system at {WAT}2017}.
\newblock In \emph{Proceedings of the 4th Workshop on {A}sian Translation ({WAT}2017)}, pages 127--134, Taipei, Taiwan. Asian Federation of Natural Language Processing.

\bibitem[{Ing{\'o}lfsd{\'o}ttir et~al.(2023)Ing{\'o}lfsd{\'o}ttir, Ragnarsson, J{\'o}nsson, Simonarson, Thorsteinsson, and Sn{\ae}bjarnarson}]{ingólfsdóttir2023bytelevelgrammaticalerrorcorrection}
Svanhv{\'\i}t~Lilja Ing{\'o}lfsd{\'o}ttir, Petur Ragnarsson, Haukur J{\'o}nsson, Haukur Simonarson, Vilhjalmur Thorsteinsson, and V{\'e}steinn Sn{\ae}bjarnarson. 2023.
\newblock \href {https://doi.org/10.18653/v1/2023.acl-long.402} {Byte-level grammatical error correction using synthetic and curated corpora}.
\newblock In \emph{Proceedings of the 61st Annual Meeting of the Association for Computational Linguistics (Volume 1: Long Papers)}, pages 7299--7316, Toronto, Canada. Association for Computational Linguistics.

\bibitem[{Jon and Bojar(2023)}]{jon-bojar-2023-breeding}
Josef Jon and Ond{\v{r}}ej Bojar. 2023.
\newblock \href {https://doi.org/10.18653/v1/2023.acl-long.122} {Breeding machine translations: Evolutionary approach to survive and thrive in the world of automated evaluation}.
\newblock In \emph{Proceedings of the 61st Annual Meeting of the Association for Computational Linguistics (Volume 1: Long Papers)}, pages 2191--2212, Toronto, Canada. Association for Computational Linguistics.

\bibitem[{Jon et~al.(2023)Jon, Popel, and Bojar}]{jon-etal-2023-cuni}
Josef Jon, Martin Popel, and Ond{\v{r}}ej Bojar. 2023.
\newblock \href {https://doi.org/10.18653/v1/2023.wmt-1.8} {{CUNI} at {WMT}23 general translation task: {MT} and a genetic algorithm}.
\newblock In \emph{Proceedings of the Eighth Conference on Machine Translation}, pages 119--127, Singapore. Association for Computational Linguistics.

\bibitem[{Jones et~al.(2023)Jones, Caswell, Firat, and Saxena}]{jones-etal-2023-gatitos}
Alexander Jones, Isaac Caswell, Orhan Firat, and Ishank Saxena. 2023.
\newblock \href {https://doi.org/10.18653/v1/2023.emnlp-main.26} {{GATITOS}: Using a new multilingual lexicon for low-resource machine translation}.
\newblock In \emph{Proceedings of the 2023 Conference on Empirical Methods in Natural Language Processing}, pages 371--405, Singapore. Association for Computational Linguistics.

\bibitem[{J{\'o}nsson et~al.(2022)J{\'o}nsson, S{\'{\i}}monarson, Ragnarsson, Ing{\'o}lfsd{\'o}ttir, {\TH}orsteinsson, and Sn{\ae}bjarnarson}]{20.500.12537/260}
Haukur~P{\'a}ll J{\'o}nsson, Haukur~Barri S{\'{\i}}monarson, P{\'e}tur~Orri Ragnarsson, Svanhv{\'{\i}}t~Lilja Ing{\'o}lfsd{\'o}ttir, Vilhj{\'a}lmur {\TH}orsteinsson, and V{\'e}steinn Sn{\ae}bjarnarson. 2022.
\newblock \href {http://hdl.handle.net/20.500.12537/260} {Long context synthetic translation pairs for english and icelandic (22.09)}.
\newblock {CLARIN}-{IS}.

\bibitem[{J{\'{o}}nsson et~al.(2020)J{\'{o}}nsson, S{\'{\i}}monarson, Sn{\ae}bjarnarson, Steingr{\'{\i}}msson, and Loftsson}]{DBLP:conf/tsd/JonssonSSSL20}
Haukur~P{\'{a}}ll J{\'{o}}nsson, Haukur~Barri S{\'{\i}}monarson, V{\'{e}}steinn Sn{\ae}bjarnarson, Stein{\th}{\'{o}}r Steingr{\'{\i}}msson, and Hrafn Loftsson. 2020.
\newblock \href {https://doi.org/10.1007/978-3-030-58323-1\_10} {{Experimenting with Different Machine Translation Models in Medium-Resource Settings}}.
\newblock In \emph{Text, Speech, and Dialogue - 23rd International Conference, {TSD} 2020, Brno, Czech Republic, September 8-11, 2020, Proceedings}, volume 12284 of \emph{Lecture Notes in Computer Science}, pages 95--103. Springer.

\bibitem[{Joulin et~al.(2016)Joulin, Grave, Bojanowski, and Mikolov}]{joulin2016bag}
Armand Joulin, Edouard Grave, Piotr Bojanowski, and Tomas Mikolov. 2016.
\newblock Bag of tricks for efficient text classification.
\newblock \emph{arXiv preprint arXiv:1607.01759}.

\bibitem[{Juraska et~al.(2023)Juraska, Finkelstein, Deutsch, Siddhant, Mirzazadeh, and Freitag}]{juraska-etal-2023-metricx}
Juraj Juraska, Mara Finkelstein, Daniel Deutsch, Aditya Siddhant, Mehdi Mirzazadeh, and Markus Freitag. 2023.
\newblock \href {https://doi.org/10.18653/v1/2023.wmt-1.63} {{M}etric{X}-23: The {G}oogle submission to the {WMT} 2023 metrics shared task}.
\newblock In \emph{Proceedings of the Eighth Conference on Machine Translation}, pages 756--767, Singapore. Association for Computational Linguistics.

\bibitem[{Khayrallah and Koehn(2018)}]{khayrallah-koehn-2018-impact}
Huda Khayrallah and Philipp Koehn. 2018.
\newblock \href {https://doi.org/10.18653/v1/W18-2709} {{On the Impact of Various Types of Noise on Neural Machine Translation}}.
\newblock In \emph{Proceedings of the 2nd Workshop on Neural Machine Translation and Generation}, pages 74--83, Melbourne, Australia. Association for Computational Linguistics.

\bibitem[{Kocmi et~al.(2024)Kocmi, Avramidis, Bawden, Bojar, Dvorkovich, Federmann, Fishel, Freitag, Gowda, Grundkiewicz, Haddow, Karpinska, Koehn, Marie, Murray, Nagata, Popel, Popovic, Shmatova, Steingrímsson, and Zouhar}]{kocmi2024preliminarywmt24rankinggeneral}
Tom Kocmi, Eleftherios Avramidis, Rachel Bawden, Ondrej Bojar, Anton Dvorkovich, Christian Federmann, Mark Fishel, Markus Freitag, Thamme Gowda, Roman Grundkiewicz, Barry Haddow, Marzena Karpinska, Philipp Koehn, Benjamin Marie, Kenton Murray, Masaaki Nagata, Martin Popel, Maja Popovic, Mariya Shmatova, Steinþór Steingrímsson, and Vilém Zouhar. 2024.
\newblock \href {https://arxiv.org/abs/2407.19884} {{Preliminary WMT24 Ranking of General MT Systems and LLMs}}.
\newblock \emph{ArXiv}, abs/2407.19884.

\bibitem[{Nakatani(2010)}]{nakatani2010langdetect}
Shuyo Nakatani. 2010.
\newblock \href {https://github.com/shuyo/language-detection} {Language detection library for java}.

\bibitem[{Olteanu et~al.(2006)Olteanu, Suriyentrakorn, and Moldovan}]{olteanu-etal-2006-language}
Marian Olteanu, Pasin Suriyentrakorn, and Dan Moldovan. 2006.
\newblock \href {https://aclanthology.org/W06-3122} {Language models and reranking for machine translation}.
\newblock In \emph{Proceedings on the Workshop on Statistical Machine Translation}, pages 150--153, New York City. Association for Computational Linguistics.

\bibitem[{Popovi{\'c}(2015)}]{popovic-2015-chrf}
Maja Popovi{\'c}. 2015.
\newblock \href {https://doi.org/10.18653/v1/W15-3049} {chr{F}: character n-gram {F}-score for automatic {MT} evaluation}.
\newblock In \emph{Proceedings of the Tenth Workshop on Statistical Machine Translation}, pages 392--395, Lisbon, Portugal. Association for Computational Linguistics.

\bibitem[{Radford et~al.(2019)Radford, Wu, Amodei, Amodei, Clark, Brundage, and Sutskever}]{radford2019better}
Alec Radford, Jeffrey Wu, Dario Amodei, Daniela Amodei, Jack Clark, Miles Brundage, and Ilya Sutskever. 2019.
\newblock Better language models and their implications.
\newblock \emph{OpenAI blog}, 1(2).

\bibitem[{Ram{\'\i}rez-S{\'a}nchez et~al.(2020)Ram{\'\i}rez-S{\'a}nchez, Zaragoza-Bernabeu, Ba{\~n}{\'o}n, and Rojas}]{prompsit:2020:EAMT}
Gema Ram{\'\i}rez-S{\'a}nchez, Jaume Zaragoza-Bernabeu, Marta Ba{\~n}{\'o}n, and Sergio~Ortiz Rojas. 2020.
\newblock \href {https://aclanthology.org/2020.eamt-1.31} {{Bifixer and Bicleaner: two open-source tools to clean your parallel data.}}
\newblock In \emph{Proceedings of the 22nd Annual Conference of the European Association for Machine Translation}, pages 291--298, Lisboa, Portugal. European Association for Machine Translation.

\bibitem[{Rei et~al.(2023)Rei, Guerreiro, Pombal, van Stigt, Treviso, Coheur, C.~de Souza, and Martins}]{rei-etal-2023-scaling}
Ricardo Rei, Nuno~M. Guerreiro, Jos{\'e} Pombal, Daan van Stigt, Marcos Treviso, Luisa Coheur, Jos{\'e}~G. C.~de Souza, and Andr{\'e} Martins. 2023.
\newblock \href {https://doi.org/10.18653/v1/2023.wmt-1.73} {Scaling up {C}omet{K}iwi: Unbabel-{IST} 2023 submission for the quality estimation shared task}.
\newblock In \emph{Proceedings of the Eighth Conference on Machine Translation}, pages 841--848, Singapore. Association for Computational Linguistics.

\bibitem[{Rei et~al.(2022)Rei, Treviso, Guerreiro, Zerva, Farinha, Maroti, C.~de Souza, Glushkova, Alves, Coheur, Lavie, and Martins}]{rei-etal-2022-cometkiwi}
Ricardo Rei, Marcos Treviso, Nuno~M. Guerreiro, Chrysoula Zerva, Ana~C Farinha, Christine Maroti, Jos{\'e}~G. C.~de Souza, Taisiya Glushkova, Duarte Alves, Luisa Coheur, Alon Lavie, and Andr{\'e} F.~T. Martins. 2022.
\newblock \href {https://aclanthology.org/2022.wmt-1.60} {{C}omet{K}iwi: {IST}-unbabel 2022 submission for the quality estimation shared task}.
\newblock In \emph{Proceedings of the Seventh Conference on Machine Translation (WMT)}, pages 634--645, Abu Dhabi, United Arab Emirates (Hybrid). Association for Computational Linguistics.

\bibitem[{S{\'a}nchez-Cartagena et~al.(2018)S{\'a}nchez-Cartagena, Ba{\~n}{\'o}n, Ortiz-Rojas, and Ram{\'\i}rez}]{sanchez-cartagena-etal-2018-prompsits}
V{\'\i}ctor~M. S{\'a}nchez-Cartagena, Marta Ba{\~n}{\'o}n, Sergio Ortiz-Rojas, and Gema Ram{\'\i}rez. 2018.
\newblock \href {https://doi.org/10.18653/v1/W18-6488} {{Prompsit{'}s submission to {WMT} 2018 Parallel Corpus Filtering shared task}}.
\newblock In \emph{Proceedings of the Third Conference on Machine Translation: Shared Task Papers}, pages 955--962, Belgium, Brussels. Association for Computational Linguistics.

\bibitem[{Schwenk et~al.(2021)Schwenk, Wenzek, Edunov, Grave, Joulin, and Fan}]{schwenk-etal-2021-ccmatrix}
Holger Schwenk, Guillaume Wenzek, Sergey Edunov, Edouard Grave, Armand Joulin, and Angela Fan. 2021.
\newblock \href {https://doi.org/10.18653/v1/2021.acl-long.507} {{CCM}atrix: Mining billions of high-quality parallel sentences on the web}.
\newblock In \emph{Proceedings of the 59th Annual Meeting of the Association for Computational Linguistics and the 11th International Joint Conference on Natural Language Processing (Volume 1: Long Papers)}, pages 6490--6500, Online. Association for Computational Linguistics.

\bibitem[{Stahl(2024)}]{stahl2024lingua}
Peter~M. Stahl. 2024.
\newblock Lingua - an accurate natural language detection library for short and mixed-language text.
\newblock \url{https://github.com/pemistahl/lingua-py}.
\newblock Accessed: 2024-08-21.

\bibitem[{Steingr{\'\i}msson et~al.(2018)Steingr{\'\i}msson, Helgad{\'o}ttir, R{\"o}gnvaldsson, Barkarson, and Gu{\dh}nason}]{steingrimsson_2018}
Stein{\th}{\'o}r Steingr{\'\i}msson, Sigrún Helgad{\'o}ttir, Eir{\'\i}kur R{\"o}gnvaldsson, Starka{\dh}ur Barkarson, and J{\'o}n Gu{\dh}nason. 2018.
\newblock \href {https://aclanthology.org/L18-1690} {{Risamálheild: A Very Large Icelandic Text Corpus}}.
\newblock In \emph{Proceedings of the Eleventh International Conference on Language Resources and Evaluation}, LREC 2018, pages 4361--4366, Miyazaki, Japan.

\bibitem[{Steingr{\'\i}msson et~al.(2023)Steingr{\'\i}msson, Loftsson, and Way}]{steingrimsson-etal-2023-filtering}
Stein{\th}{\'o}r Steingr{\'\i}msson, Hrafn Loftsson, and Andy Way. 2023.
\newblock \href {https://aclanthology.org/2023.nodalida-1.58} {Filtering matters: Experiments in filtering training sets for machine translation}.
\newblock In \emph{Proceedings of the 24th Nordic Conference on Computational Linguistics (NoDaLiDa)}, pages 588--600, T{\'o}rshavn, Faroe Islands. University of Tartu Library.

\bibitem[{Steingr{\'\i}msson et~al.(2021)Steingr{\'\i}msson, Lohar, Loftsson, and Way}]{steingrimsson-etal-2021-effective}
Stein{\th}{\'o}r Steingr{\'\i}msson, Pintu Lohar, Hrafn Loftsson, and Andy Way. 2021.
\newblock \href {https://aclanthology.org/2021.bucc-1.3} {{Effective Bitext Extraction From Comparable Corpora Using a Combination of Three Different Approaches}}.
\newblock In \emph{Proceedings of the 14th Workshop on Building and Using Comparable Corpora (BUCC 2021)}, pages 8--17, Online (Virtual Mode). INCOMA Ltd.

\bibitem[{Steingrímsson(2023)}]{Steingrimsson2023Phd}
Steinþór Steingrímsson. 2023.
\newblock \emph{Effectively compiling parallel corpora for machine translation in resource-scarce conditions}.
\newblock Ph.D. thesis, Reykjavik University.

\bibitem[{Steingrímsson et~al.(2023)Steingrímsson, Loftsson, and Way}]{steingrimsson-etal-2023-sentalign}
Steinþór Steingrímsson, Hrafn Loftsson, and Andy Way. 2023.
\newblock \href {https://doi.org/10.18653/v1/2023.emnlp-demo.22} {{S}ent{A}lign: Accurate and scalable sentence alignment}.
\newblock In \emph{Proceedings of the 2023 Conference on Empirical Methods in Natural Language Processing: System Demonstrations}, pages 256--263, Singapore. Association for Computational Linguistics.

\bibitem[{Tiedemann and Thottingal(2020)}]{tiedemann-thottingal-2020-opus}
J{\"o}rg Tiedemann and Santhosh Thottingal. 2020.
\newblock \href {https://aclanthology.org/2020.eamt-1.61} {{OPUS}-{MT} {--} building open translation services for the world}.
\newblock In \emph{Proceedings of the 22nd Annual Conference of the European Association for Machine Translation}, pages 479--480, Lisboa, Portugal. European Association for Machine Translation.

\bibitem[{Toral et~al.(2023)Toral, Cranenburgh, and Nutters}]{toraletal2023}
Antonio Toral, Andreas Cranenburgh, and Tia Nutters. 2023.
\newblock \href {https://doi.org/10.4324/9781003357391-3} {Literary-adapted machine translation in a well-resourced language pair}.
\newblock In Andrew Rothwell, Andy Way, and Roy Youdale, editors, \emph{Computer-Assisted Literary Translation}, pages 27--52. Routledge.

\bibitem[{Vamvas and Sennrich(2022)}]{https://doi.org/10.48550/arxiv.2204.13692}
Jannis Vamvas and Rico Sennrich. 2022.
\newblock \href {https://aclanthology.org/2022.findings-emnlp.15} {{NMTS}core: A multilingual analysis of translation-based text similarity measures}.
\newblock In \emph{Findings of the Association for Computational Linguistics: EMNLP 2022}, pages 198--213, Abu Dhabi, United Arab Emirates.

\bibitem[{Vaswani et~al.(2017)Vaswani, Shazeer, Parmar, Uszkoreit, Jones, Gomez, Kaiser, and Polosukhin}]{vaswani2017}
Ashish Vaswani, Noam Shazeer, Niki Parmar, Jakob Uszkoreit, Llion Jones, Aidan~N Gomez, Łukasz Kaiser, and Illia Polosukhin. 2017.
\newblock \href {http://papers.nips.cc/paper/7181-attention-is-all-you-need.pdf} {{Attention is All you Need}}.
\newblock In \emph{Advances in Neural Information Processing Systems 30 (NIPS 2017)}, pages 5999--6009, Long Beach, California.

\bibitem[{Wang et~al.(2007)Wang, Stolcke, and Zheng}]{4430102}
Wen Wang, Andreas Stolcke, and Jing Zheng. 2007.
\newblock \href {https://doi.org/10.1109/ASRU.2007.4430102} {Reranking machine translation hypotheses with structured and web-based language models}.
\newblock In \emph{2007 IEEE Workshop on Automatic Speech Recognition \& Understanding (ASRU)}, pages 159--164.

\bibitem[{Wormer(2024)}]{wormer2024franc}
Titus Wormer. 2024.
\newblock Franc - a natural language detection library.
\newblock \url{https://github.com/wooorm/franc}.
\newblock Accessed: 2024-08-21.

\bibitem[{Xu et~al.(2024{\natexlab{a}})Xu, Kim, Sharaf, and Awadalla}]{xu2024a}
Haoran Xu, Young~Jin Kim, Amr Sharaf, and Hany~Hassan Awadalla. 2024{\natexlab{a}}.
\newblock \href {https://openreview.net/forum?id=farT6XXntP} {{A Paradigm Shift in Machine Translation: Boosting Translation Performance of Large Language Models}}.
\newblock In \emph{The Twelfth International Conference on Learning Representations}.

\bibitem[{Xu et~al.(2024{\natexlab{b}})Xu, Sharaf, Chen, Tan, Shen, Durme, Murray, and Kim}]{xu2024contrastivepreferenceoptimizationpushing}
Haoran Xu, Amr Sharaf, Yunmo Chen, Weiting Tan, Lingfeng Shen, Benjamin~Van Durme, Kenton Murray, and Young~Jin Kim. 2024{\natexlab{b}}.
\newblock \href {https://api.semanticscholar.org/CorpusID:267028540} {{Contrastive Preference Optimization: Pushing the Boundaries of LLM Performance in Machine Translation}}.
\newblock \emph{ArXiv}, abs/2401.08417.

\bibitem[{Xue et~al.(2022)Xue, Barua, Constant, Al-Rfou, Narang, Kale, Roberts, and Raffel}]{xue2022byt5tokenfreefuturepretrained}
Linting Xue, Aditya Barua, Noah Constant, Rami Al-Rfou, Sharan Narang, Mihir Kale, Adam Roberts, and Colin Raffel. 2022.
\newblock \href {https://doi.org/10.1162/tacl_a_00461} {{B}y{T}5: Towards a token-free future with pre-trained byte-to-byte models}.
\newblock \emph{Transactions of the Association for Computational Linguistics}, 10:291--306.

\end{thebibliography}
\bibliographystyle{acl_natbib}

\appendix

\section{OPUS Texts}
\label{sec:appendix_opus}

The parallel texts we sourced from the OPUS catalog are listed in this section. The format of the list is as follows:
\vspace{2mm}\\
\noindent\textit{Index}. \textbf{Name}; \textit{version}; sentence pairs
\vspace{2mm}\\
For brevity, the \textit{ELRC} parallel text names are abbreviated after the first entry in the list, with the \textit{ditto} symbol (`\textquotedbl') replacing the `ELRC' part of the name.
\vspace{2mm}\\
\noindent 1. \textbf{CCAligned}; \textit{v1}; \hfill1,192,542 \\
\noindent 2. \textbf{CCMatrix}; \textit{v1}; \hfill8,723,145 \\
\noindent 3. \textbf{ECDC}; \textit{v2016-03-16}; \hfill2,512 \\
\noindent 4. \textbf{ELRC-2718-EMEA}; \textit{v1}; \hfill542,624 \\
\noindent 5. \textbf{\textquotedbl-3206-antibiotic}; \textit{v1}; \hfill816 \\
\noindent 6. \textbf{\textquotedbl-4295-www.malfong.is}; \textit{v1}; \hfill12,634 \\
\noindent 7. \textbf{\textquotedbl-4324-Government\_Offices\_I}; \textit{v1}; \hfill18,185 \\
\noindent 8. \textbf{\textquotedbl-4327-Government\_Offices\_I}; \textit{v1}; \hfill36,290 \\
\noindent 9. \textbf{\textquotedbl-4334-Rkiskaup\_2020}; \textit{v1}; \hfill10,236 \\
\noindent 10. \textbf{\textquotedbl-4338-University\_Iceland}; \textit{v1}; \hfill10,164 \\
\noindent 11. \textbf{\textquotedbl-502-Icelandic\_Financial\_}; \textit{v1}; \hfill1,525 \\
\noindent 12. \textbf{\textquotedbl-504-www.iceida.is}; \textit{v1}; \hfill1,055 \\
\noindent 13. \textbf{\textquotedbl-505-www.pfs.is}; \textit{v1}; \hfill2,866 \\
\noindent 14. \textbf{\textquotedbl-506-www.lanamal.is}; \textit{v1}; \hfill1,140 \\
\noindent 15. \textbf{\textquotedbl-5067-SciPar}; \textit{v1}; \hfill110,831 \\
\noindent 16. \textbf{\textquotedbl-508-Tilde\_Statistics\_Ice}; \textit{v1}; \hfill2,427 \\
\noindent 17. \textbf{\textquotedbl-509-Gallery\_Iceland}; \textit{v1}; \hfill577 \\
\noindent 18. \textbf{\textquotedbl-510-Harpa\_Reykjavik\_Conc}; \textit{v1}; \hfill1,197 \\
\noindent 19. \textbf{\textquotedbl-511-bokmenntaborgin\_is}; \textit{v1}; \hfill330 \\
\noindent 20. \textbf{\textquotedbl-516-Icelandic\_Medicines}; \textit{v1}; \hfill711 \\
\noindent 21. \textbf{\textquotedbl-517-Icelandic\_Directorat}; \textit{v1}; \hfill1,536 \\
\noindent 22. \textbf{\textquotedbl-597-www.nordisketax.net}; \textit{v1}; \hfill1,065 \\
\noindent 23. \textbf{\textquotedbl-718-Statistics\_Iceland}; \textit{v1}; \hfill2,361 \\
\noindent 24. \textbf{\textquotedbl-728-www.norden.org}; \textit{v1}; \hfill41,073 \\
\noindent 25. \textbf{\textquotedbl-EMEA}; \textit{v1}; \hfill542,624 \\
\noindent 26. \textbf{\textquotedbl-antibiotic}; \textit{v1}; \hfill816 \\
\noindent 27. \textbf{\textquotedbl-www.norden.org}; \textit{v1}; \hfill41,073 \\
\noindent 28. \textbf{\textquotedbl-www.nordisketax.net}; \textit{v1}; \hfill1,065 \\
\noindent 29. \textbf{EUbookshop}; \textit{v2}; \hfill9,783 \\
\noindent 30. \textbf{GNOME}; \textit{v1}; \hfill28,776 \\
\noindent 31. \textbf{HPLT}; \textit{v1}; \hfill2,148,876 \\
\noindent 32. \textbf{KDE4}; \textit{v2}; \hfill98,989 \\
\noindent 33. \textbf{MaCoCu}; \textit{v2}; \hfill267,366 \\
\noindent 34. \textbf{MultiCCAligned}; \textit{v1}; \hfill1,192,537 \\
\noindent 35. \textbf{MultiHPLT}; \textit{v1}; \hfill2,148,855 \\
\noindent 36. \textbf{MultiMaCoCu}; \textit{v2}; \hfill267,366 \\
\noindent 37. \textbf{MultiParaCrawl}; \textit{v7.1}; \hfill2,392,423 \\
\noindent 38. \textbf{NLLB}; \textit{v1}; \hfill8,723,145 \\
\noindent 39. \textbf{OpenSubtitles}; \textit{v1}; \hfill7,138 \\
\noindent 40. \textbf{OpenSubtitles}; \textit{v2016}; \hfill1,359,224 \\
\noindent 41. \textbf{OpenSubtitles}; \textit{v2018}; \hfill1,569,189 \\
\noindent 42. \textbf{ParIce}; \textit{v1}; \hfill2,097,022 \\
\noindent 43. \textbf{ParaCrawl}; \textit{v7.1}; \hfill2,392,422 \\
\noindent 44. \textbf{ParaCrawl}; \textit{v8}; \hfill5,724,373 \\
\noindent 45. \textbf{ParaCrawl}; \textit{v9}; \hfill2,967,579 \\
\noindent 46. \textbf{QED}; \textit{v2.0a}; \hfill27,611 \\
\noindent 47. \textbf{TED2020}; \textit{v1}; \hfill2,430 \\
\noindent 48. \textbf{Tatoeba}; \textit{v2}; \hfill8,139 \\
\noindent 49. \textbf{Tatoeba}; \textit{v20190709}; \hfill9,436 \\
\noindent 50. \textbf{Tatoeba}; \textit{v2020-05-31}; \hfill9,438 \\
\noindent 51. \textbf{Tatoeba}; \textit{v2020-11-09}; \hfill9,440 \\
\noindent 52. \textbf{Tatoeba}; \textit{v2021-03-10}; \hfill9,443 \\
\noindent 53. \textbf{Tatoeba}; \textit{v2021-07-22}; \hfill9,443 \\
\noindent 54. \textbf{Tatoeba}; \textit{v2022-03-03}; \hfill9,522 \\
\noindent 55. \textbf{Tatoeba}; \textit{v2023-04-12}; \hfill9,600 \\
\noindent 56. \textbf{TildeMODEL}; \textit{v2018}; \hfill420,712 \\
\noindent 57. \textbf{Ubuntu}; \textit{v14.10}; \hfill2,155 \\
\noindent 58. \textbf{WikiMatrix}; \textit{v1}; \hfill85,992 \\
\noindent 59. \textbf{WikiTitles}; \textit{v3}; \hfill50,176 \\
\noindent 60. \textbf{XLEnt}; \textit{v1}; \hfill962,661 \\
\noindent 61. \textbf{XLEnt}; \textit{v1.1}; \hfill962,661 \\
\noindent 62. \textbf{XLEnt}; \textit{v1.2}; \hfill962,661 \\
\noindent 63. \textbf{bible-uedin}; \textit{v1}; \hfill62,163 \\
\noindent 64. \textbf{wikimedia}; \textit{v20190628}; \hfill581 \\
\noindent 65. \textbf{wikimedia}; \textit{v20210402}; \hfill2,625 \\
\noindent 66. \textbf{wikimedia}; \textit{v20230407}; \hfill4,471 \\

\section{Filtering steps}
\label{sec:appendix_filter}

\noindent\textbf{Filter 1. Sentence length}\\
Sentences should contain at minimum four characters and at maximum 150 characters.\vspace{2mm}

\noindent\textbf{Filter 2. High inter-pair content overlap}\\
Sentence pairs where the content of the source and target sentences are highly similar should be removed from the dataset.\vspace{2mm}

\noindent\textbf{Filter 3. Character symbol filtering}\\
All characters in the English and Icelandic alphabets (along with punctuation and numbers) designated as a set of allowed characters. Sentences containing less than 60\% of these characters removed from the data and all characters outside the allowed set removed from the remaining sentences.\footnote{This is the last filtering step that inherently modifies the content inside individual sentences.}\vspace{2mm}

\noindent\textbf{Filter 4. LaBSE scoring}\\
We use score each sentence pair using LaBSE \cite{feng-etal-2022-language} and remove all sentences with a score lower than 0.8\footnote{This is a higher cutoff than the original LaBSE authors suggest to use, but our experiments suggets it better suits our data.}. \vspace{2mm}

\noindent\textbf{Filter 5. Language detection}\\
We use various language detection software to gauge whether both the source and target sentences are in the correct language. The software we used was \textit{fasttext} \cite{joulin2016bag}, \textit{franc} \cite{wormer2024franc}, \textit{lingua} \cite{stahl2024lingua} and \textit{langdetect} \cite{nakatani2010langdetect}.\vspace{2mm}

\noindent\textbf{Filter 6. Similar dataset pairs}\\ 
As a safeguard, we remove any duplicate entries of our dataset if, for any reason, there remain duplicate instances after the previous filters. In our final experiment, this was rendered redundant, but was required in previous iterations and may prove useful in future iterations.\vspace{2mm}

\noindent\textbf{Filter 7. Near-duplicate dataset pairs}\\
Sentences are compared by removing content-specific words that are likely proper names and dates, etc., and comparing the remainder.\vspace{2mm}

\noindent\textbf{Filter 8. Likely machine-translated target sentences}\\
A GPT-2 \cite{radford2019better} classifier is used to evaluate whether a given target sentence is machine-translated, based on a 10.000 sentence hand-evaluated reference set. If this is true for the target sentence, that pair is removed from the dataset.\vspace{2mm}

\noindent\textbf{Filter 9. Existing datasets}\\
As a final safeguard check, we remove any sentence pair that we already have on file in other datasets, as touched on in section \ref{sec:opus}.\vspace{2mm}

\noindent\textbf{Filter 10. NMTScore cross-likelyhood 0.4}\\
Finally, we use a translation cross-likelyhood NMTScore \cite{https://doi.org/10.48550/arxiv.2204.13692} to determine the translation quality of a given sentence pair. This step is computationally heavy and was therefore saved for last. Our experiments suggest that 0.4 is a suitable cutoff for our dataset.

\end{document}